\documentclass[journal,twoside,web]{ieeecolor}
\usepackage{xcolor}
\usepackage{jsen}
\usepackage{cite}
\usepackage{amsmath,amssymb,amsfonts}
\usepackage{algorithmic}
\usepackage{graphicx}
\usepackage{textcomp}
\usepackage{wrapfig}
\usepackage{siunitx}

\def\BibTeX{{\rm B\kern-.05em{\sc i\kern-.025em b}\kern-.08em
    T\kern-.1667em\lower.7ex\hbox{E}\kern-.125emX}}
\definecolor{abstractbg}{rgb}{0.89804,0.94510,0.83137}
\begin{document}

\title{External Photoreflective Tactile Sensing Based on Surface Deformation Measurement}

\author{Seiichi Yamamoto, Hiroki Ishizuka, Takumi Kawasetsu, ~\IEEEmembership{Member, IEEE}, Koh Hosoda,~\IEEEmembership{Senior Member, IEEE}, Takayuki Kameoka, Kango Yanagida, Takato Horii, ~\IEEEmembership{Member, IEEE}, \textcolor{black}{Yusuke Sakaue}, Sei Ikeda,~\IEEEmembership{Member, IEEE}, Osamu Oshiro
\thanks{Manuscript received: October 6, 2024; Revised: January 9, 2025; Accepted: March 17, 2025.}
\thanks{This study was supported by JST PRESTO Grant Number JPMJPR22S2, JST CREST Grant Number JPMJCR2555, and JSPS KAKENHI Grant numbers 22H01447 and 23H01379, Japan.}
\thanks{S. Yamamoto, H. Ishizuka, K. Yanagida, T. Horii, \textcolor{black}{Y. Sakaue}, S. Ikeda, and O. Oshiro are with the Graduate School of Engineering Science, The University of Osaka, Toyonaka 560-0043, Japan (e-mail: yama19991015@icloud.com; ishizuka@bpe.es.osaka-u.ac.jp; k.yanagida@rlg.sys.es.osaka-u.ac.jp; takato@sys.es.osaka-u.ac.jp; \textcolor{black}{sakaue.yuusuke.es@osaka-u.ac.jp}; ikeda.sei.jp@ieee.org; osamu.oshiro.es@osaka-u.ac.jp).}
\thanks{T. Kawasetsu and K. Hosoda are with the Department of Mechanical Engineering and Science,
Kyoto University, Kyoto 606-8501, Japan (e-mail: kawasetsu.takumi.2f@kyoto-u.ac.jp; hosoda.koh.7p@kyoto-u.ac.jp).}
\thanks{T. Kameoka is with the Faculty of Design Department of Media Design, Kyushu University, Fukuoka 819-0395, Japan (e-mail: kameoka@design.kyushu-u.ac.jp).}
}

\IEEEtitleabstractindextext{%
\fcolorbox{abstractbg}{abstractbg}{%
\begin{minipage}{\textwidth}%
\begin{wrapfigure}[12]{r}{3in}%
\includegraphics[width=3in]{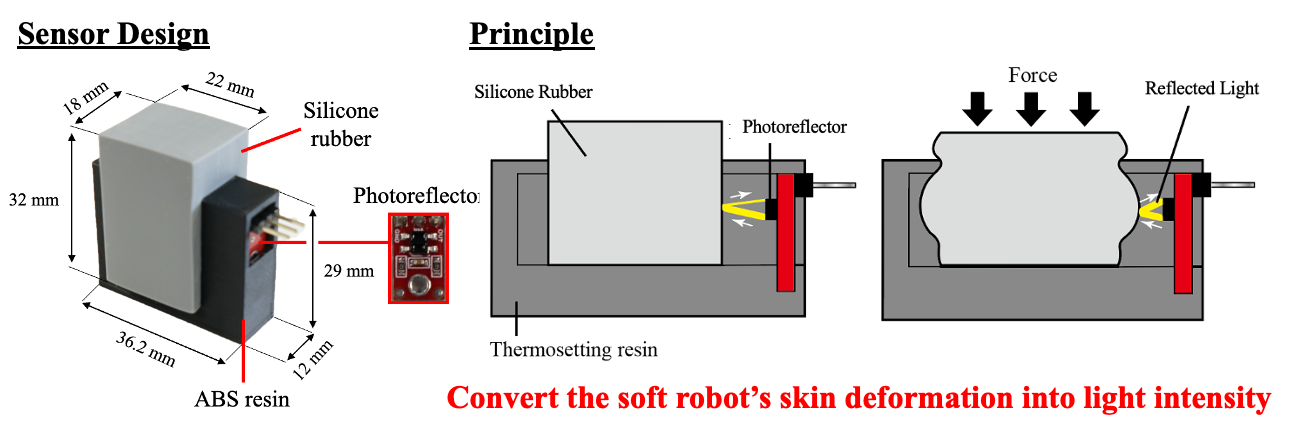}%
\end{wrapfigure}%
\begin{abstract}
We present a tactile-sensing method enabled by the mechanical compliance of soft robots: an externally attachable photoreflective module reads surface deformation of silicone skin to estimate contact force \textcolor{black}{while (i) locating the sensing module off the contact interface, (ii) keeping the skin free of embedded transducers/wiring, and (iii) enabling rapid replacement and reconfiguration of the module}. 
Locating the sensor off the contact interface reduces damage risk, \textcolor{black}{and the non-embedded architecture preserves softness} \textcolor{black}{while simplifying fabrication and maintenance through a plug-in, replaceable module}. We first characterize the optical sensing element and the compliant skin, then determine the design of a prototype tactile sensor. Compression experiments validate the approach, exhibiting a monotonic force--output relationship consistent with theory, low hysteresis, high repeatability over repeated cycles, and small response lag over 0.1--10~mm/s indentation speeds. We further demonstrate integration on a soft robotic gripper, where the module reliably detects grasp events. Compared with liquid-filled or wire-embedded tactile skins, the proposed modular ``add-on'' architecture enhances durability, reduces wiring complexity, \textcolor{black}{and supports easy redeployment across diverse robot geometries without disassembling the robot skin}. Because the sensing principle reads skin-strain patterns, it also suggests extensions to other somatosensory cues---such as joint-angle or actuator-state estimation---from surface deformation. Overall, leveraging surface compliance with an external optical module provides a practical and robust route to equip soft robots with force perception while preserving structural flexibility and manufacturability, paving the way for robotic applications and safe human--robot collaboration.
\end{abstract}

\end{minipage}}}

\maketitle

\section{INTRODUCTION}
\label{sec:introduction}
\IEEEPARstart{R}{obots} with inherent softness are expected to be applied in a wide range of fields, including healthcare, human-robot collaboration, and disaster response, owing to their safety when working near humans and their adaptability to various environments~\cite{katarobomondai, gendairobot, gendairobot2}. To realize soft robots, it is essential to employ innovative mechanisms and materials that impart softness to their structures. One promising approach is to fabricate the constituent components of a robot using soft materials. For example, recent efforts have been directed toward the development of soft sensors~\cite{Nakadegawa2017, Ohashi2023}, actuators~\cite{Hirai2021, Tanaka2025}, and computing devices~\cite{GarradSoftMatterComputer2019, Hashimoto2023}. Among these, soft tactile sensors are particularly crucial because they enable robots to perceive and interact with their surroundings, similar to humans. These sensors allow robots to gather, interpret, and utilize information regarding external objects and environments. Therefore, the development of soft sensors is the key to developing robots that are both safe and highly adaptable.

Soft tactile sensors acquire tactile information by converting the deformation of their transducer elements caused by the mechanical deformation of the sensor body in response to external stimuli. \textcolor{black}{Soft tactile sensors are an essential technology for enabling soft robots to sense tactile information without compromising their inherent compliance.} The transduction mechanisms can be broadly categorized into several types: piezoelectric sensors that generate a voltage upon mechanical stress~\cite{atudenkoukasiki,atudenkoukasikiondo}; magnetic sensors that detect variations in magnetic fields due to the deformation of soft components~\cite{zikisiki}; resistive sensors that exploit changes in the electrical resistance of materials or structures under force~\cite{teikousiki1,teikousiki2}; capacitive sensors that measure changes in capacitance resulting from deformation~\cite{seidenyouryousiki1,seidenyouryousiki2,seidenyouryousiki3}; and optical sensors that detect variations in light intensity due to the deformation of optical fibers~\cite{hikarifiber1,hikarifiber2,hikarifiber3}. 

\textcolor{black}{In general, implementing such tactile sensors on soft robots requires placing or embedding them near the robot surface where external forces are applied~\cite{umekomi1,umekomi2,umekomi3}. This configuration maximizes the deformation of the transducer under load and can therefore improve sensitivity. However, integrating tactile sensors into soft robots often complicates the fabrication process by introducing additional sensor fabrication steps~\cite{umekomi1,umekomi2,umekomi3}. Moreover, repeated contact or exposure to excessive forces can cause damage to the sensing elements and transducers, as well as wire breakage~\cite{dansen}. Because repairing these sensors is not straightforward, it may become necessary to rebuild the soft robot itself. For long-term use of soft robots, external-force sensing that combines durability and maintainability is required. From this perspective, key requirements for realizing such sensing include: (i) locating the sensing modules off the contact surface to reduce the risk of damage; (ii) simplifying the structure and manufacturing process by avoiding the embedding of sensing components---such as conductors, discrete elements, waveguides, or fluidic channels---within the soft skin (contact surface); and (iii) enabling easy replacement and reconfiguration of the sensor module to ensure high maintainability. 
}

One approach to address the aforementioned challenges in tactile sensing is to design sensors that leverage the inherent softness of materials. In the human body, the skin stretches in response to physical movements~\cite{Mia2025}. For example, it has been observed that bending the knee or making facial expressions induces distinct patterns of skin strain. It is hypothesized that the spatially distributed mechanoreceptors within the skin convert these strain patterns into action potentials, which are then transmitted to the brain~\cite{johansson2004}. According to this principle, it is plausible that when a soft robot moves or experiences a force on its surface, a unique strain pattern emerges across its compliant exterior. 

\begin{figure}[!t]
\centerline{\includegraphics[width=1\columnwidth]{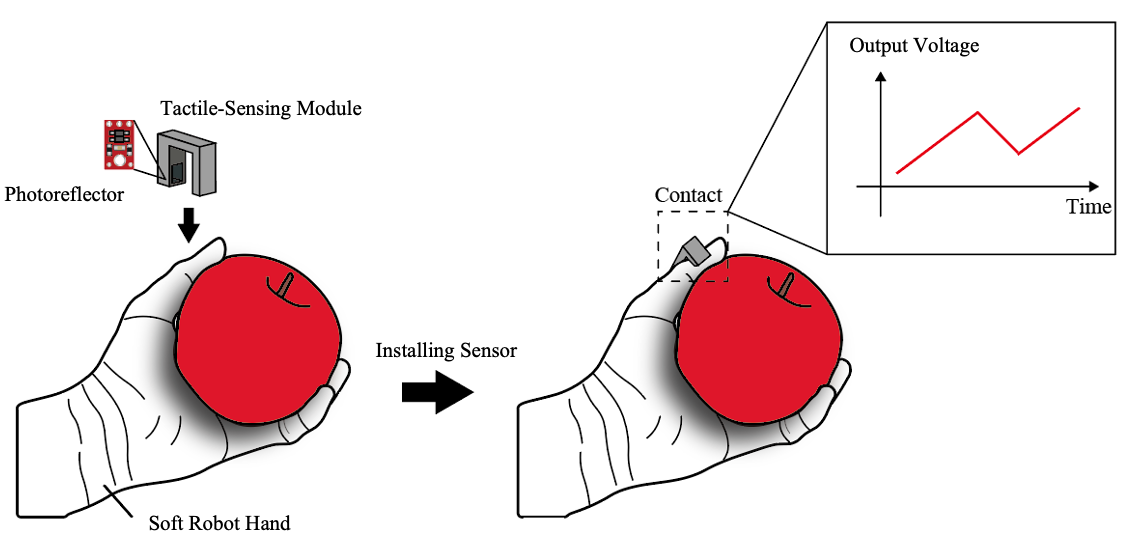}}
\caption{Layout example of the proposed tactile-sensing method. When a soft robot performs a tactile interaction with an object, a deformation distribution arises on its skin surface. By measuring this deformation using a tactile-sensing module attached externally to the skin, force information can be obtained without placing a transducer directly at the contact interface.}
\label{fig1}
\end{figure}

\textcolor{black}{In this study, we propose a method for acquiring tactile information---specifically, force sensing---without embedding sensing elements inside a soft body. In the proposed approach, deformations occurring on the robot's surface are detected using an externally attached tactile-sensing module, similar to a wearable device (Fig.~\ref{fig1}). A force applied at the contact surface of a soft body also induces deformation in distant regions of the soft body. By appropriately measuring this remote deformation, contact forces can be estimated without placing sensors near the contact interface.
The proposed approach satisfies requirement (i). Moreover, because it relies on measuring deformation at a location away from where the force is applied, it does not require incorporating any special structures within the soft robot skin; it only requires attaching the external module which can be easily attached and detached. In prior studies that satisfy requirement (i), however, a liquid-sealing process inside the soft robot skin was required~\cite{soter2019, usui2021, hamaguchi2021}, and thus requirement (ii) was not met. In contrast, our approach also satisfies requirements (ii) and (iii). This distinction is crucial for long-term experimentation and deployment, where surface layers are frequently exposed to repeated contacts, abrasion, and occasional overload, making repair and reconfiguration inevitable.
}

Previous studies have demonstrated techniques for estimating the shape of a soft robotic arm by attaching inertial measurement units to its exterior~\cite{Martin2022}. However, research on estimating a robot's state using externally mounted sensors remains limited. To the best of our knowledge, no studies have explicitly focused on force estimation in this manner.
Our method is based on the principle of converting the deformation of a soft body into changes in light intensity. This principle was also applied to human skin-based force sensing~\cite{nakatani2011, Saito2019}. 
Unlike human skin, which exhibits large individual variability~\cite{kaneko2025}, silicone rubber has stable material properties. Therefore, when applied to soft robots, the sensor performance can be more readily validated through simulation and more easily optimized through design.
Although force sensors utilizing optical reflection have been proposed~\cite{ohmura2006, cAndrea2021, hayase2023}, these designs are typically integrated into the internal structure of a robot. In this study, we adopted an external mounting structure for soft robots, enabling both ease of fabrication and improved sensor maintainability while allowing accurate force estimation. There have been limited studies on measuring a soft robot's body state without embedding sensors in the robot body~\cite{Martin2022}, and the concept presented in this work can contribute to advancing soft robot design and system integration. 
\textcolor{black}{We position the proposed method as a modular add-on for soft-robot skins, emphasizing robustness, maintainability, and ease of integration across diverse geometries.}
The contributions of this study are as follows.

\begin{itemize}
\item We investigated the design requirements suitable for implementing the proposed method.
\item We fabricated a prototype sensor with the simplest structure that still simulates actual usage and evaluated its sensing characteristics.
\item We demonstrated an application by integrating the tactile-sensing module into a soft robot and evaluated its capability to detect forces induced by external loads and contacts.
\end{itemize}

\section{SENSOR DESIGN}
In this study, we developed a prototype of a tactile sensor that estimates contact force by converting the deformation of a soft body induced by an applied force into a change in light reflectance. To guide the design, we evaluated the characteristics of the employed photoreflector and the silicone rubber individually. According to these evaluations, we constructed a model for the sensor design and simulation.

\subsection{Sensing Principle}
\begin{figure}[!t]
\centerline{\includegraphics[width=1\columnwidth]{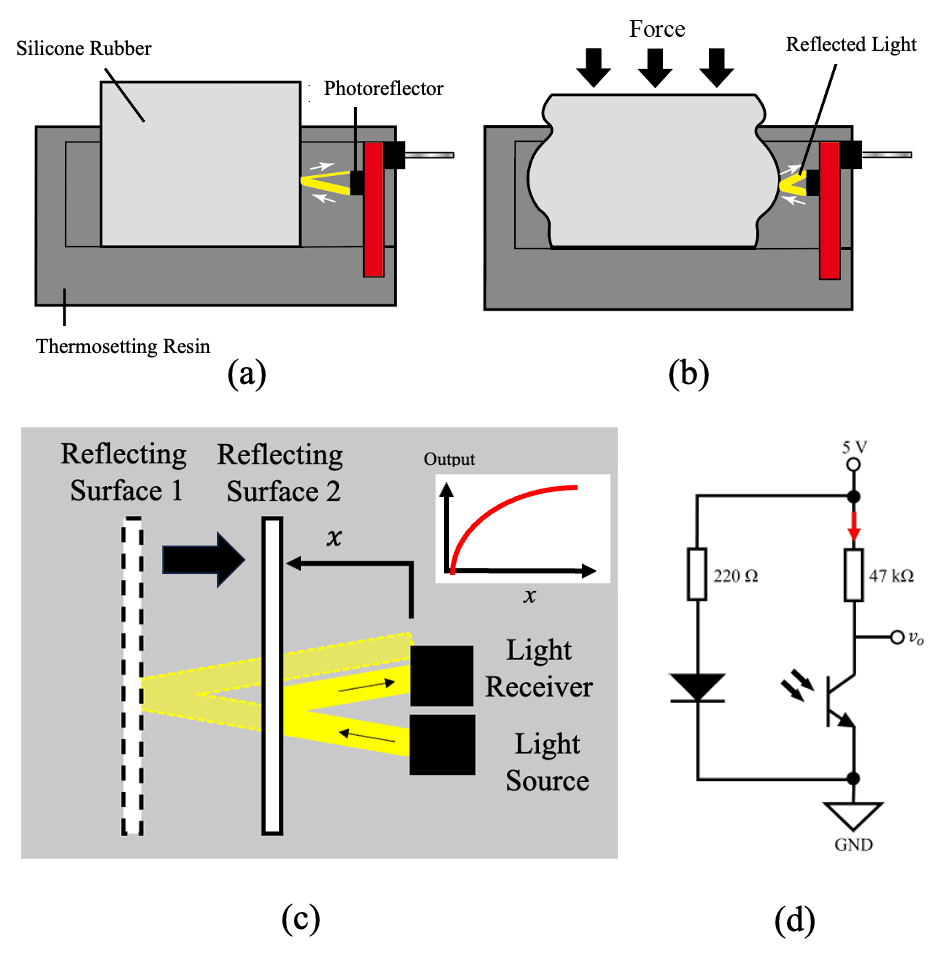}}
\caption{Principle of the proposed tactile-sensing method. (a) A tactile-sensing module containing a photoreflector, which emits light and measures the intensity of the reflected light, is attached to a soft body such as silicone rubber. (b, c) When an external force deforms the soft body, the distance between the sensing element and the reflective surface changes. This change in distance causes a change in the amount of reflected light. (d) The variation in the amount of the reflected light causes the variation in output voltage of the photoreflector. The applied external force can then be estimated from the variation in output voltage corresponding to the variation in received light intensity.}
\label{fig2}
\end{figure}

The fundamental concept of the proposed method is to estimate the net normal forces by measuring the deformation of a soft body caused by such forces using an externally mounted tactile-sensing module. This approach enables force measurement using an externally attached module without embedding sensing elements inside the soft robot. To verify the feasibility of the proposed method, we implemented the approach in a simple structural configuration and evaluated its sensing characteristics. Inspired by the works of Nakatani et al.\ and Saito et al.~\cite{nakatani2011, Saito2019}, a tactile sensor was constructed by fitting a deformation-measurement module to a soft silicone rubber structure (Figs.~\ref{fig2}(a) and (b)). The module incorporated a photoreflector that emitted light and converted the intensity of the reflected light into a voltage signal. Although alternative sensing approaches could be adopted, the photoreflector was selected in this study because of its simplicity and ability to convert the lateral deformation of the soft structure into an output signal.

The module can be easily detached, enabling straightforward replacement in case of damage or adaptation to different robot geometries. The photoreflector is positioned to direct light toward the silicone rubber and receive the reflected light. When an external force compresses the silicone rubber, its incompressibility causes it to shrink in the longitudinal direction and expand laterally. Consequently, the distance between the photoreflector and the silicone rubber decreases, increasing the amount of reflected light (Fig.~\ref{fig2}(c)). As shown in Fig.~\ref{fig2}(d), due to the internal circuit of the photoreflector, the output voltage decreases as the intensity of the reflected light increases. This change in the reflected light intensity produces a corresponding change in the photoreflector's output voltage, which can then be used to estimate the applied force.
\textcolor{black}{The proposed method in this study employs a single photoreflector for sensing; therefore, it can only estimate the force applied to the entire surface, rather than the spatial distribution of forces. In this study, the calibration and evaluation are performed under a controlled loading condition using a flat indenter that compresses the whole top surface, which promotes an approximately uniform deformation of the silicone rubber. We acknowledge that localized or off-center contacts may lead to position-dependent responses; this limitation and possible extensions are discussed in Sec. IV.}

\subsection{Sensing Element}
The characteristics of a photoreflector (QTR-1A, POLOLU) were evaluated to guide the sensor design. The output of a photoreflector depends on the characteristics of its internal components, the distance to the target object, and the target's reflectance. Therefore, to design the sensor, it is necessary to investigate the relationship between the photoreflector output and the distance between the photoreflector and the silicone rubber. The amount of light reflected from the target is determined by factors such as the material, surface geometry, and color of the target. In this study, we focused on color, as it is an easily adjustable parameter during the design process. By altering the color of the target, its reflectance can be modified, allowing us to explore the conditions under which the photoreflector's output characteristics are suitable for sensor design.

\begin{figure}[!t]
\centerline{\includegraphics[width=1\columnwidth]{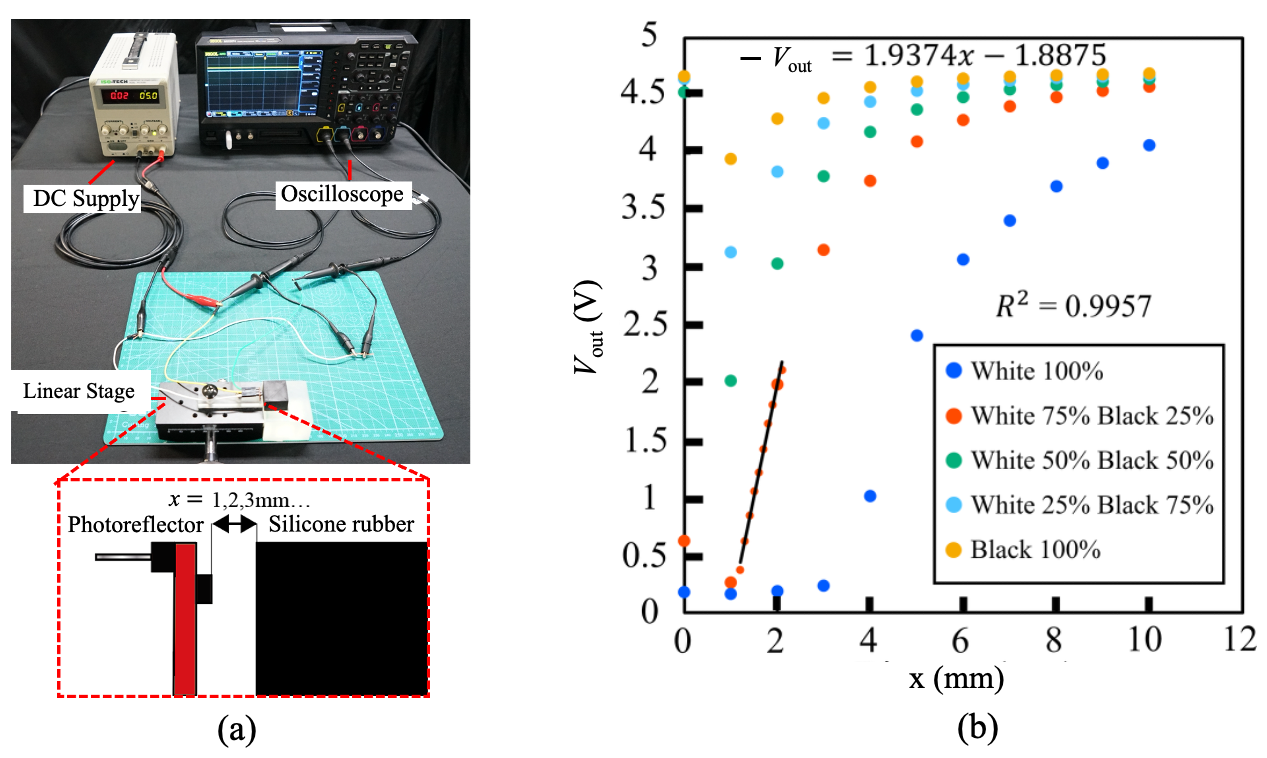}}
\caption{The output voltage of the photoreflector was evaluated as a function of the color of the silicone rubber and the intensity of the reflected light. (a) Experimental setup: A linear stage was used to vary the distance between the colored silicone rubber and the photoreflector, and the resulting changes in output voltage were measured. (b) The output voltage increased with increasing distance. The output characteristics also varied depending on the color of the silicone rubber. According to the experimental result, a mixture ratio of 75\% white and 25\% black silicone rubber was adopted in this study.}
\label{fig3}
\end{figure}

The experimental setup is illustrated in Fig.~\ref{fig3}(a). The distance between the photoreflector and the silicone rubber was defined as $x$ (mm). The silicone rubber was mounted on a linear stage (LS-912, Chuo Precision Industrial Co., Ltd.) and brought into contact with a photoreflector ($x = 0$~mm). The linear stage was then moved to increase the distance (0--10~mm), and the photoreflector output was recorded using an oscilloscope (MSO5074, RIGOL Technologies). Trials were conducted in a dark room to eliminate the influence of ambient light.

The effect of the silicone rubber color was also examined. Silicone rubber was mixed with white and black pigments (Silc-Pig and Smooth-On) at various ratios. The total silicone weight was 8.2~g, and the total pigment weight was 0.2~g. Five samples labeled silicones 1--5 were prepared with white:black pigment ratios of 100\%:0\%, 75\%:25\%, 50\%:50\%, 25\%:75\%, and 0\%:100\%, respectively.

The relationship between the photoreflector output and the photoreflector--silicone distance is shown in Fig.~\ref{fig3}(b). The change in the output was larger for samples with a higher proportion of white pigment. This is attributed to the lower light absorption of the white silicone rubber relative to that of the black silicone rubber. A larger change in the output with respect to distance reduces the susceptibility to noise-induced errors and potentially improves sensitivity. Therefore, under the tested conditions, silicone with a high proportion of white pigment was preferable.

The proposed tactile sensor estimates the contact force according to the difference in photoreflector output. To simplify the conversion from voltage to force, the photoreflector output should be approximately linear with respect to distance. Assuming that the deformation surface of the silicone rubber is planar, the distance--voltage relationship shown in Fig.~\ref{fig3}(b) can be used to estimate output.
\textcolor{black}{The distance between the photoreflector and the silicone rubber was minimized to keep the module smaller. We evaluated the relationship and selected an initial distance of approximately 2 mm, within the 1--2~mm range where good linearity was observed. Within this range, we chose the pigmentation condition that provided the large voltage change; specifically, a mixture of 75\% white pigment and 25\% black pigment was selected.}

\subsection{Dimensions}
\begin{figure}[!t]
\centerline{\includegraphics[width=1\columnwidth]{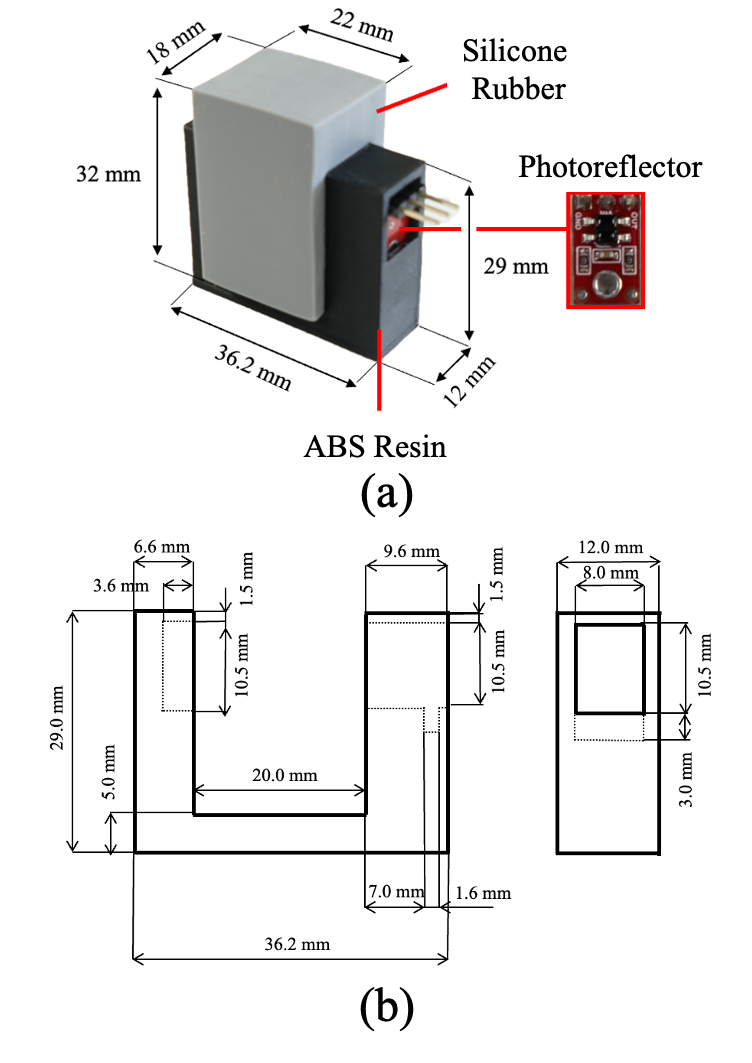}}
\caption{Prototype tactile sensor and tactile-sensing module developed in this study. (a) Photograph of the actual sensor and its dimensions. (b) Illustration of the tactile-sensing module and its dimensions. The module is simply fixed to the silicone rubber, representing a straightforward implementation of tactile information sensing.}
\label{fig4}
\end{figure}

The fabricated tactile sensor for the proof-of-concept is shown in Fig.~\ref{fig4}(a). A silicone rubber body (Ecoflex-0030, Smooth-On) was molded using an acrylonitrile butadiene styrene (ABS) resin mold fabricated using a 3D printer (da Vinci Jr. ProX+, XYZ Printing), and a tactile-sensing module was integrated with the silicone rubber body. The tactile-sensing module shown in Fig.~\ref{fig4}(b) had an embedded photoreflector to measure the deformation of the silicone rubber. 
\textcolor{black}{The infrared emitter in the photoreflector is driven at 17~mA (VIN=5~V), with a peak emission wavelength of 940~nm. The detector is a phototransistor.}
\textcolor{black}{Its dark current is $\leq$100~nA, and the coupled on-state collector current is 0.10--0.40~mA under the continuous forward current of 20~mA and the saturation voltage of 5~V (mirror at the distance of 1~mm).}

\textcolor{black}{To minimize the effect of ambient light reaching the photoreflector, the ABS resin was colored black, giving it a high light absorption rate. The width of the silicone rubber body was 22~mm. A dent slightly narrower than the silicone rubber body was formed; thus, the elastic reaction force of the silicone rubber helped secure the module in place when inserted, making it difficult to detach. The distance between the photoreflector and the silicone rubber surface was approximately 2~mm. This structure allows the module to be easily fixed to a silicone rubber body without adhesives or double-sided tape. When a contact force is applied, the distance between the photoreflector and the silicone rubber decreases, reducing the output, as shown in Fig.~\ref{fig3}(b). With this structure, the proposed sensing method can be implemented on any soft material regardless of its shape by simply inserting and securing the module.}

\begin{figure}[!t]
\centerline{\includegraphics[width=1\columnwidth]{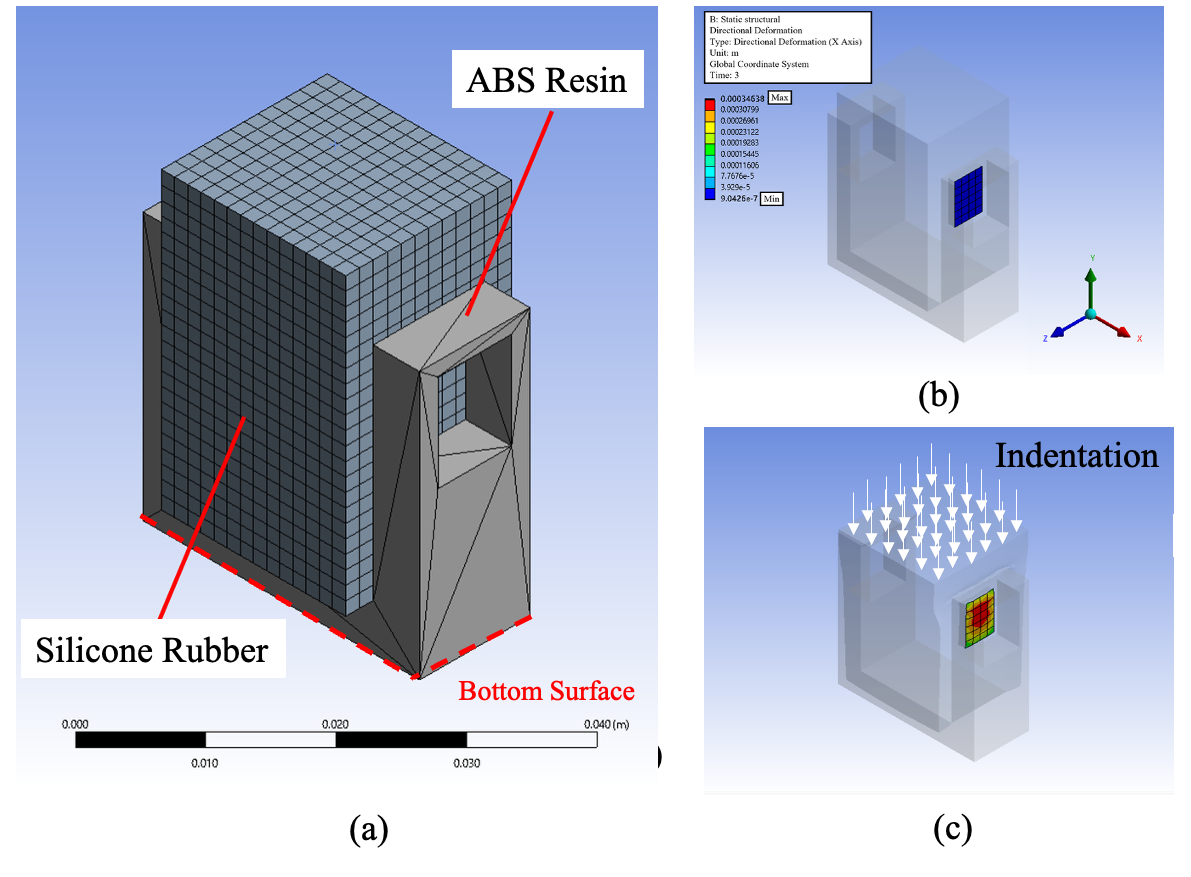}}
\caption{A finite-element model was used for the analysis. (a) The model consists of a structure representing the silicone rubber and the transducer. (b) Before indentation is applied. (c) After indentation is applied. Applying the indentation causes deformation on the side surface of the silicone rubber. This deformation value is used to calculate the theoretical change in output voltage.}
\label{fig5}
\end{figure}

\subsection{Simulation Model}
\textcolor{black}{A finite-element analysis software (ANSYS, Ansys Inc.) was used to construct a model for deriving the theoretical output of the photoreflector in response to contact forces. A finite-element model with the same dimensions as the fabricated tactile sensor was prepared (Fig.~\ref{fig5}(a)). A static analysis was performed. The silicone rubber was modeled as a hyperelastic material using the parameters presented in Table~\ref{tab1}~\cite{niu2022} using the Mooney--Rivlin model. For the ABS resin module attached to the outer surface of the silicone rubber, the parameters listed in Table~\ref{tab2} were applied, as provided in the software library. Hexahedral elements were used for the silicone rubber, and tetrahedral elements for the module. The finite-element model was meshed with 7005 nodes and 1713 elements, and the bottom surface of the sensor was fixed as the boundary condition. The entire top surface of the silicone rubber body was displaced downward by 3~mm. When an indentation was applied to the top of the sensor, deformation occurred in the side region where the photoreflector was placed (Fig.~\ref{fig5}(c)).}

\begin{table}
\caption{Parameters of silicone rubber~\cite{niu2022}.}
\label{tab1}
\setlength{\tabcolsep}{3pt}
\begin{tabular}{|p{25pt}|p{75pt}|p{115pt}|}
\hline
Symbol& Definition& Values \\
\hline
$C_{01}$ & Material constant & $1.218\times10^{-2}\ \si{MPa}$\\
$C_{10}$ & Material constant & $-3.335\times10^{-5}\ \si{MPa}$\\
\hline
\end{tabular}
\end{table}

\begin{table}
\caption{Parameters of ABS resin at \SI{23}{\degreeCelsius}.}
\label{tab2}
\setlength{\tabcolsep}{3pt}
\begin{tabular}{|p{25pt}|p{125pt}|p{75pt}|}
\hline
Symbol & Definition & Values \\
\hline
$E$     & Young's modulus           & $1.628\times 10^{9}~\si{Pa}$ \\
$\nu$   & Poisson's ratio           & $4.089\times 10^{-1}$ \\
$K$     & Bulk modulus              & $2.978\times 10^{9}~\si{Pa}$ \\
$G$     & Shear modulus             & $5.778\times 10^{8}~\si{Pa}$ \\
$\rho$  & Density                   & $1.030\times 10^{3}~\si{kg/m^{3}}$ \\
$\sigma$& Tensile yield strength    & $2.744\times 10^{7}~\si{Pa}$ \\
$R_m$   & Ultimate tensile strength & $3.626\times 10^{7}~\si{Pa}$ \\
\hline
\end{tabular}
\end{table}

\textcolor{black}{Furthermore, the average displacement of the elements closest to the photoreflector was used to calculate the distance $x$ (mm) between the photoreflector and the silicone rubber. By using the linear approximation shown in Fig.~\ref{fig3}(b), the change in the photoreflector output $V_{\mathrm{out}}$ can be calculated according to
\begin{equation}
V_{\mathrm{out}} = 1.9374x - 1.8875.
\label{eq1}
\end{equation}
This linear approximation describes how the photoreflector output $V_{\mathrm{out}}$ varies with distance $x$. Combined with the applied indentation, this equation allows the output of the tactile-sensing module to be expressed as a function of indentation.}

\section{EXPERIMENT}
\textcolor{black}{Experiments were conducted using the proposed tactile sensor to estimate the contact force from variations in the output. }

\subsection{Experimental Setup}
\begin{figure}[!t]
\centerline{\includegraphics[width=1\columnwidth]{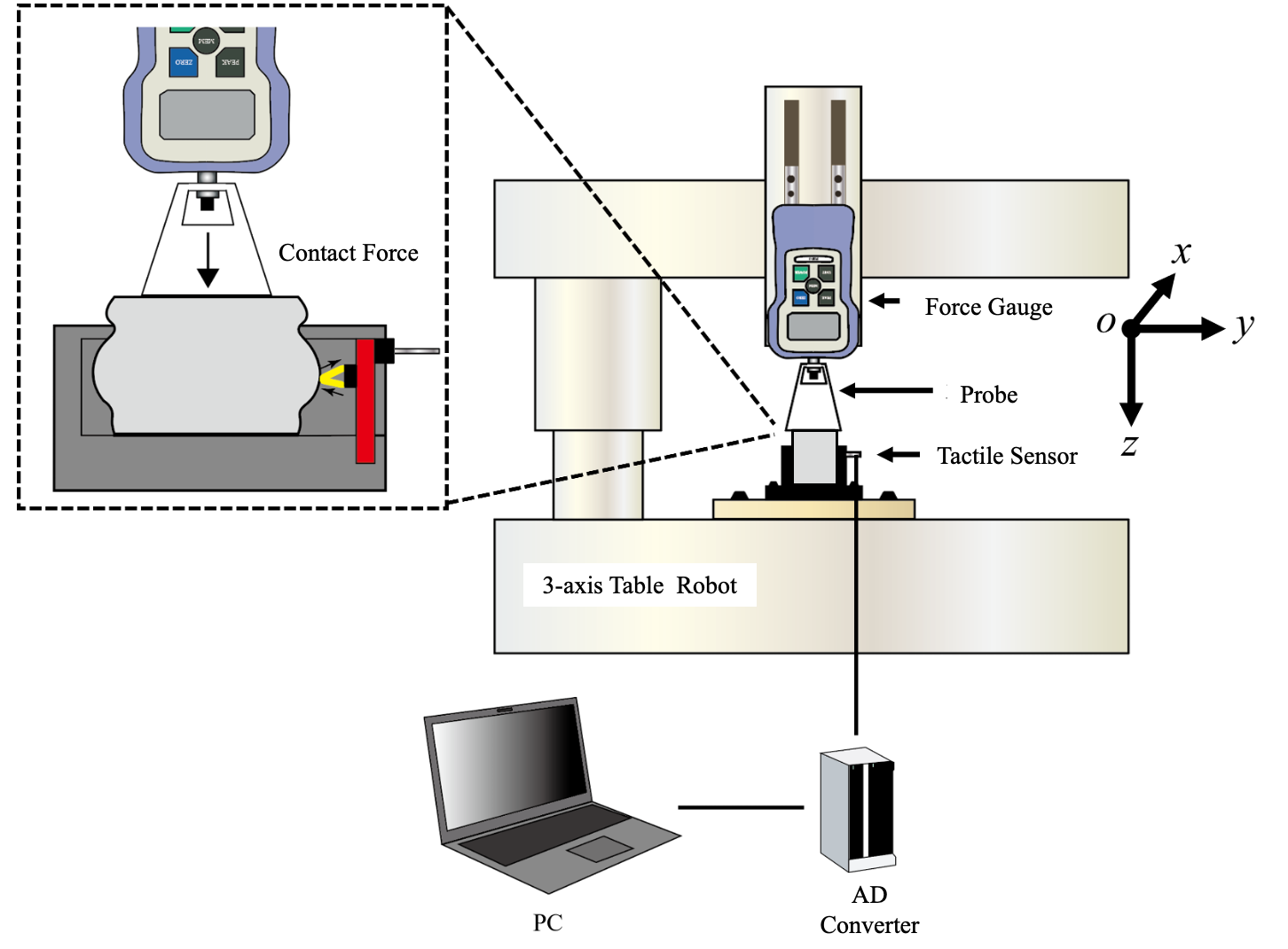}}
\caption{Configuration of the experimental setup. The setup consisted of a positioning table robot, a force gauge, a data logger, and a laptop PC. Force was applied to the sensor using the positioning table robot, while the outputs from both the force gauge and the sensor were recorded by the data logger and monitored on the PC.}
\label{fig6}
\end{figure}

\textcolor{black}{Fig.~\ref{fig6} shows the experimental setup for evaluating the sensing response to contact forces applied in the pressing direction. The fabricated tactile sensor was attached to a three-axis automatic positioning table robot (TTA-C3, IAI). A digital force gauge (FGP-5, Nidec) equipped with a 40~mm $\times$ 40~mm square probe contact area was mounted on the $z$-axis slider of the positioning table robot. A contact force was applied to the tactile sensor by moving the $z$-axis slider. The outputs of the tactile-sensing module and the digital force gauge were acquired using an analog-to-digital (AD) converter (NR-500, KEYENCE) at a sampling frequency of 1~kHz. To plot the contact force data, the noise originating from the vibrations of the positioning table's motor was removed by applying a 9-point moving average filter.} \textcolor{black}{In the experiment to evaluate the effect of temperature, a different table robot (ICSB3-BA1MB1L-WA-30AQ-25AQ-10AQB-T2-3L-CT-CT) and a force sensor (USLG25-10N, Tec Gihan) with equivalent functionality were used. In this experiment, another AD converter (AI-1608AY-USB, CONTEC) was used. The sampling rate was set to 10 Hz, and a moving-average filter with a window size of 4 was applied.}

\textcolor{black}{The output of the tactile-sensing module was defined as the value obtained by inverting the output voltage of the photoreflector. This processing compensates for the inverse correlation between the photoreflector signal and applied force, enabling direct interpretation of the module output as a proxy for contact force.}

\begin{figure*}[!t]
    \centering
    \vspace{3mm}
    \includegraphics[width=\textwidth]{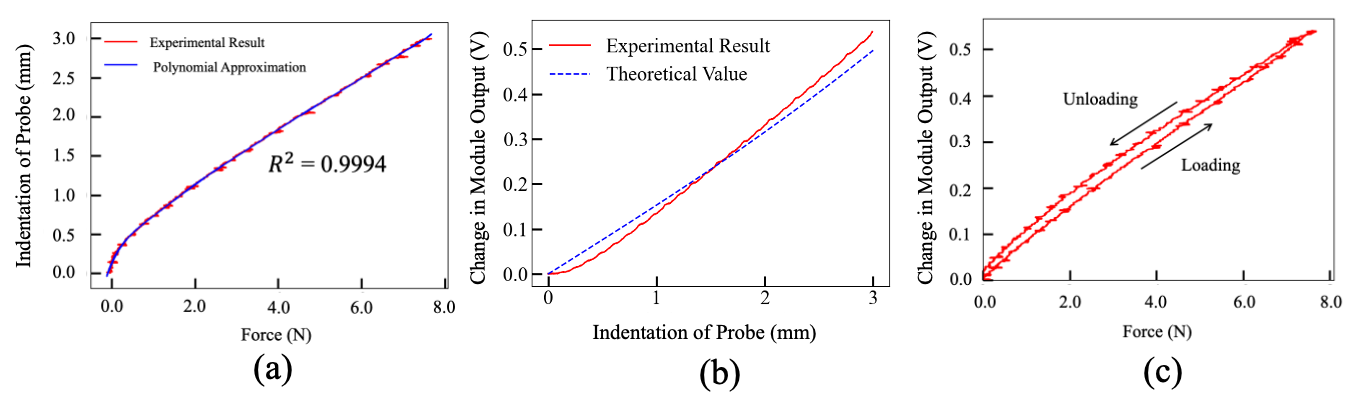}
    \vspace{-5mm}
    \caption{Evaluation results of the tactile-sensing module's response to applied force. The vertical axis represents the change in the module output (inverted photoreflector output, see Section~III-A). (a) Relationship between the applied force and indentation of the silicone rubber. A monotonic increasing relationship was observed between force and indentation. (b) Relationship between the indentation and the output of the tactile-sensing module, along with the theoretical values. A monotonic increasing relationship was observed between indentation and output. The close agreement between the theoretical and measured values indicates that the tactile sensor was well modeled. (c) Relationship between the applied force and the output of the tactile-sensing module. Owing to the low viscoelasticity of the silicone rubber, favorable hysteresis characteristics were obtained.}
    \vspace{-6mm}
    \label{fig7}
\end{figure*}

\begin{figure*}[!t]
    \centering
    \vspace{3mm}
    \includegraphics[width=\textwidth]{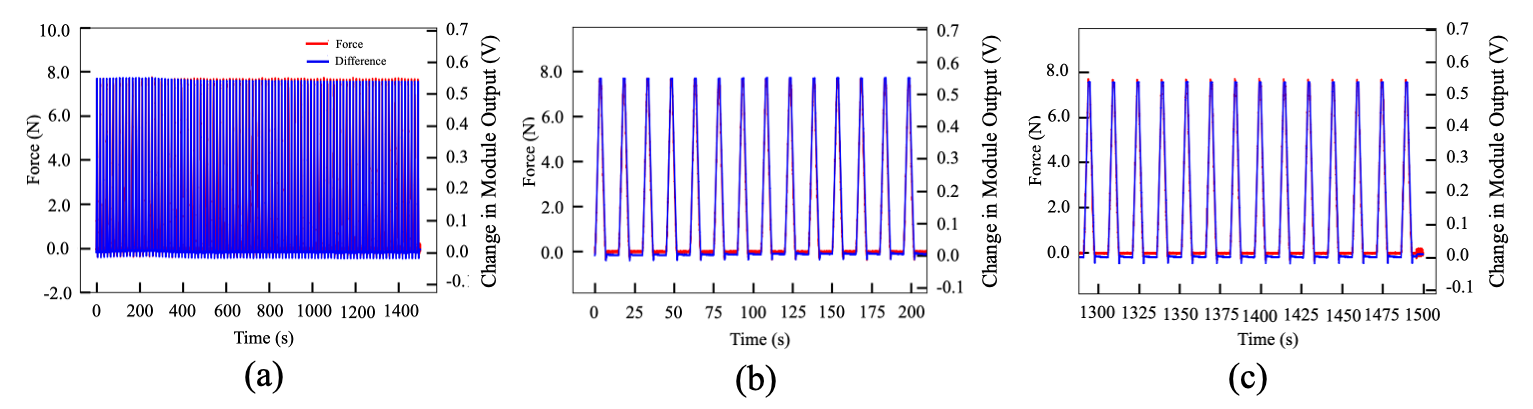}
    \vspace{-5mm}
    \caption{Sensor response under repeated loading. The vertical axis represents the change in the module output (inverted photoreflector output, see Section~III-A). (a) The output of the tactile-sensing module remained constant regardless of the number of loading cycles. (b, c) Outputs at the beginning and end of the experiment. In both cases, the voltage changes closely followed the changes in the applied force, regardless of the number of cycles, demonstrating high repeatability.}
    \vspace{-6mm}
    \label{fig8}
\end{figure*}

\begin{figure}[!t]
\centerline{\includegraphics[width=1\columnwidth]{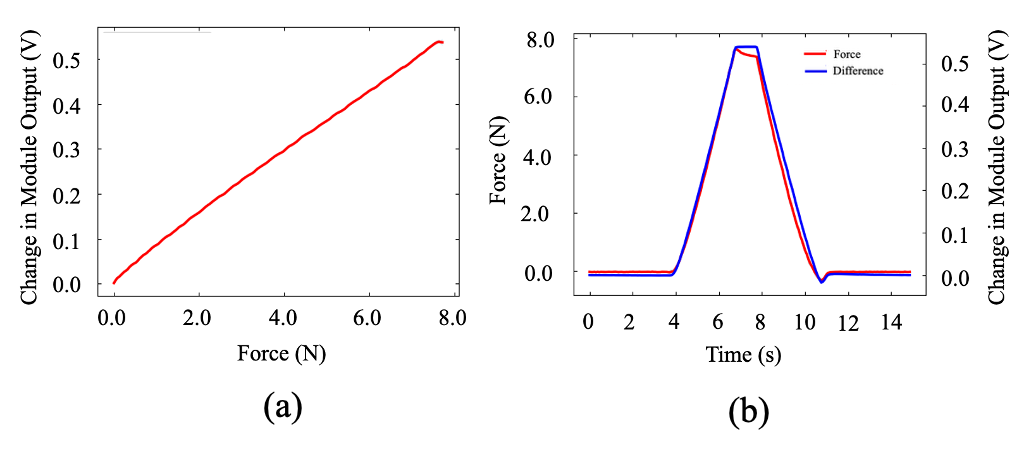}}
\caption{\textcolor{black}{Repeatability analysis results. (a) Relationship between the voltage output and the applied load obtained from 99 segmented loading trials. The mean and standard deviation are plotted, and the standard deviation is very small, indicating good repeatability. (b) Time histories of the mean voltage and force over the 98 segmented loading-unloading cycles. The standard deviation is also very small, confirming good repeatability.}}
\label{ex1}
\end{figure}

\begin{figure*}[!t]
    \centering
    \vspace{3mm}
    \includegraphics[width=\textwidth]{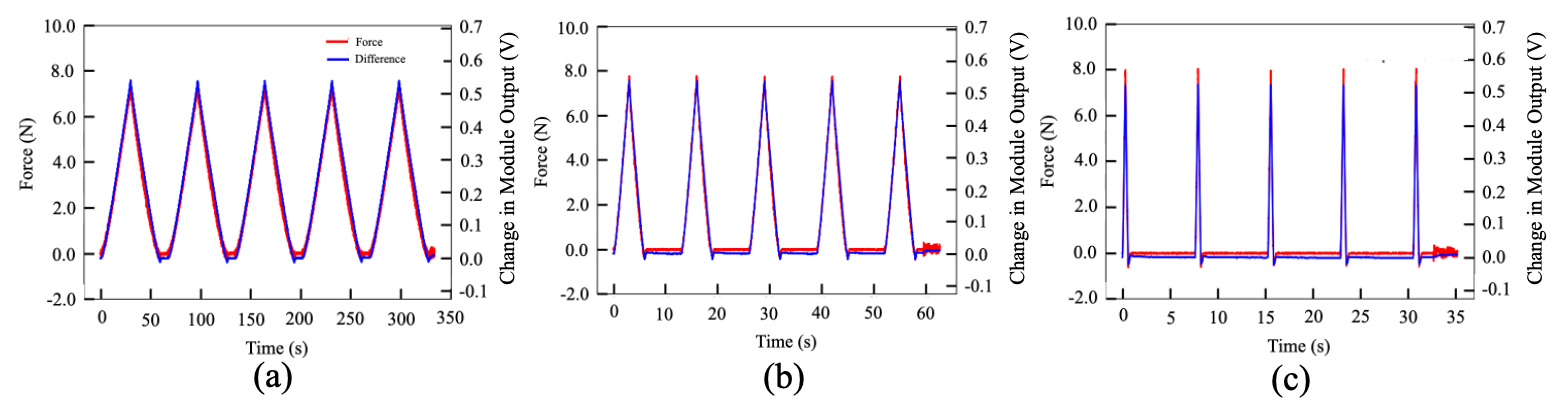}
    \vspace{-5mm}
    \caption{Sensor response under different loading and unloading speeds: (a) 0.1~mm/s, (b) 1~mm/s, and (c) 10~mm/s. The vertical axis represents the change in the module output (inverted photoreflector output, see Section~III-A). In all loading conditions, the output closely followed the applied force, indicating that the output followed the applied force, with speed-dependent lag. This behavior is attributed to the elasticity and low viscoelasticity of the silicone rubber.}
    \vspace{-6mm}
    \label{fig9}
\end{figure*}

\begin{figure*}[!t]
    \centering
    \vspace{3mm}
    \includegraphics[width=\textwidth]{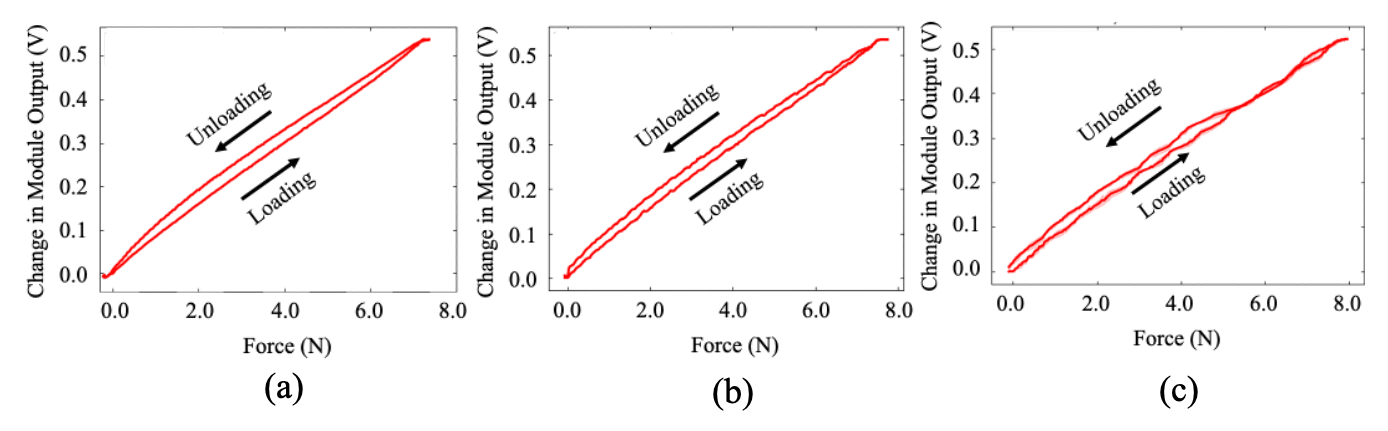}
    \vspace{-5mm}
    \caption{\textcolor{black}{Results for hysteresis: (a) 0.1~mm/s, (b) 1~mm/s, and (c) 10~mm/s. The hysteresis was derived from the results of five repeated trials. The maximum hysteresis error  was approximately 6--8\%FS.}}
    \vspace{-6mm}
    \label{ex2}
\end{figure*}

\subsection{Response to Contact Force}
 \textcolor{black}{The relationship between the applied contact force and the change in the output of the tactile-sensing module was evaluated and compared with the simulation results. Starting from the state where the probe was in contact with the sensor, the sensor was pressed vertically downward at a speed of 1~mm/s with a maximum indentation of 3~mm. During this process, the output of the tactile-sensing module and the contact force were recorded.} \textcolor{black}{This process was conducted once.}  \textcolor{black}{First, the relationship between the indentation and contact force was derived. Then, we compared the output of the tactile-sensing module with the simulation results. Finally, the relationship between the contact force and the output of the tactile-sensing module was obtained during both loading and unloading, and the hysteresis characteristics were evaluated.}

 \textcolor{black}{The relationship between the probe indentation and the contact force is shown in Fig.~\ref{fig7}(a). The proposed sensor exhibited an indentation of 3~mm when a force of approximately 7~N was applied. In the indentation range of 0--0.5~mm, the relationship between the contact force and the indentation was nonlinear. In contrast, when the probe indentation exceeded approximately 0.5~mm, a linear relationship was observed between the contact force and the indentation. These results indicate that a nonlinear relationship exists between contact force and indentation.} \textcolor{black}{This behavior is presumed to arise from the combined effects of the hyperelastic behavior of the silicone rubber, the constraint imposed by the ABS resin, and nonuniform contact in the small-indentation region.}

 \textcolor{black}{Fig.~\ref{fig7}(b) shows the relationship between the change in output of the tactile-sensing module from its initial value and the applied indentation. The finite-element analysis simulation confirmed that, in the gap around the photoreflector, the silicone rubber displaced as designed. It was confirmed that the change in the output of the tactile-sensing module increased monotonically with an increase in the indentation. The theoretical changes in output from the initial value are plotted as blue dots in Fig.~\ref{fig7}(b). For clarity, the inverted photoreflector output (as defined in Section~III-A) is used in the following plots. The measured and theoretical values agreed well overall. The sensitivity was 0.07~V/N and the dynamic range was 31.49~dB.}  

 \textcolor{black}{Although the theoretical values exhibited an almost linear change in voltage, the measured values exhibited a nonlinear trend at small contact forces. This discrepancy is attributed to the modeling of the surface. In the modeling process, the deformed side surface of the silicone rubber was assumed to be planar, which may have contributed to this difference. Although complete agreement between the theoretical and experimental results was not achieved, the differences were small, confirming that the proposed sensor was designed as intended and that the theoretical model was able to reproduce the experimental behavior.}

 \textcolor{black}{Fig.~\ref{fig7}(c) presents the relationship between the change in output of the tactile-sensing module from its initial value and the applied contact force. It was confirmed that the change in the output of the tactile-sensing module increased approximately proportionally with an increase in the contact force.} 
 \textcolor{black}{The contact force--output relationship was nearly identical during loading and unloading and the calibration of the sensor is considered feasible without additional compensation or adjustment.}

\subsection{Repeatability}
 \textcolor{black}{The repeatability of the sensing was evaluated according to the measured output of the tactile-sensing module. The probe was moved at a speed of 1~mm/s, with a maximum indentation of 3~mm.} \textcolor{black}{The loading--unloading cycles were repeated 100 times to evaluate the output repeatability}.  \textcolor{black}{A waiting time of 1~s was set between loading and unloading. The contact force and the output were recorded during the cyclic tests.}

 \textcolor{black}{Fig.~\ref{fig8}(a) presents the measured output corresponding to the contact force in the indentation direction over 100 repeated loading cycles. Because no change was observed in the maximum output, it can be concluded that the sensor exhibited high repeatability under indentation. The relationship between the output and the contact force was examined in further detail. Fig.~\ref{fig8}(b) presents the results for the first 10 loading cycles, whereas Fig.~\ref{fig8}(c) shows the results for the last 10 cycles. The output characteristics of the sensor remained unchanged after 100 loading--unloading cycles. Furthermore, the variations in the force and output were synchronized, confirming that the proposed sensor was able to capture force changes accurately. This repeatability is attributed to the properties of silicone rubber. The silicone rubber had a low viscosity; therefore, once the load was removed, it returned to its original shape, and its deformation characteristics did not change under repeated loading. }

\textcolor{black}{For 99 extracted cycles, Fig.~\ref{ex1}(a) shows the relationship between the change in output of the tactile-sensing module from its initial value and the applied contact force during loading. The mean voltage (solid line) and its standard deviation (shaded band) are plotted at each force point. The standard deviation was too small to be shown, indicating small variability and high repeatability of the response. In addition, the output change $\Delta V = V_{\mathrm{high}} - V_{\mathrm{low}}$ and the input change $\Delta F = F_{\mathrm{high}} - F_{\mathrm{low}}$ were computed from the steady-state low/high plateaus for each cycle, and the sensitivity was quantified as $S=\Delta V/\Delta F$. The output change and input change (Mean~$\pm$~SD) were $\Delta V = 0.467 \pm 0.004~\mathrm{V}$ and
$\Delta F = 6.46 \pm 0.06~\mathrm{N}$, respectively. Across the cycles, the cycle-to-cycle variability of the change and the force was small.
The sensitivity (Mean~$\pm$~SD) was $S = 0.07236 \pm 0.00037~\mathrm{V/N}$, demonstrating that the sensitivity remained nearly constant over repeated loading.
Fig.~\ref{ex1}(b) overlays the force and voltage waveforms over the 98 cycles and reports the mean (solid line) with $\pm$ SD (shaded band) as a function of time. 
The standard deviations were too small to be shown and  confirm that the temporal response of the sensor was consistent across repetitions.
Random uncertainty was quantified as the calculated standard deviation from repeated trials. Systematic uncertainty was estimated from the instrument specifications (Force gauge: $\pm$0.2\% R.C; AD converter: $\pm$0.03\%).
}

\subsection{Effect of Indentation Speed}
\textcolor{black}{The output of the tactile-sensing module, which depended on the deformation of the silicone rubber, was measured to confirm that the deformation rate followed the contact force application rate. In other words, the influence of the indentation speed on the sensing response was evaluated. The probe was moved at speeds of 0.1, 1, and 10~mm/s with a maximum indentation of 3~mm.} \textcolor{black}{The loading--unloading cycles were repeated five times.} \textcolor{black}{The contact force and output were recorded during the cyclic tests.}

\textcolor{black}{Fig.~\ref{fig9} shows the results of an experiment for evaluating the effect of the indentation speed on the output response. The output of the tactile-sensing module followed the changes in the contact force when the speed was 0.1~mm/s and 1~mm/s. As the speed increased, the measured force tended to become higher. In contrast, the maximum sensor output showed little dependence on speed; even when compared at the same applied indentation within the rising segment, the outputs were essentially consistent. This behavior is attributed to the viscoelasticity of silicone rubber, which increases the apparent stiffness under faster loading, thereby requiring a higher force for the same indentation. The phase lag between the applied load and the sensor output was 5~ms at 10~mm/s, 62~ms at 1~mm/s, and 706~ms at 0.1~mm/s. Qiu et al.\ reported an approximately 0.045~s response time to step inputs~\cite{qiu2019}, as a sensitive sensor. Although a direct comparison is difficult due to differences in stimuli (step vs.\ ramp) and metrics, we consider that at indentation speeds of $\geq 1$~mm/s, our sensor also detects force with only a practically slight delay.}

\textcolor{black}{Because high repeatability under cyclic loading was confirmed, we evaluated the hysteresis characteristics using the results from five independent trials (n~=~5). The results are shown in Fig.~\ref{ex2}. 
The values (solid line) and their standard deviations (shaded bands) are plotted. The standard deviation was too small to be shown. The maximum hysteresis obtained by computation (Mean~$\pm$~SD) was 6.50~$\pm$~0.21\%FS at 0.1~mm/s, 6.50~$\pm$~0.22\%FS at 1~mm/s, and 7.88~$\pm$~0.31\%FS at 10~mm/s. As expected, the standard deviations of the hysteresis were small. The maximum hysteresis reported for tactile sensors based on elastomeric materials is typically within 5--10\%~\cite{Wang2017, cAndrea2021, usui2021}, indicating that our sensor exhibits hysteresis characteristics comparable to existing soft tactile sensors. The tendency for the maximum hysteresis to increase with increasing loading speed is consistent with that seen in the previous study\cite{cAndrea2021}. 
These results confirm that a sensor architecture without embedded sensing elements can achieve hysteresis performance comparable to conventional sensors that embed sensing elements.}

\subsection{Effect of Temperature}
\textcolor{black}{The effect of temperature variation on the output was then evaluated. An initial pre-load of 0.6~N was applied. The probe velocity was set to 1~mm/s, and the maximum indentation depth was 3~mm. Loading and unloading were repeated 20~times. A waiting time of 1 s~was set between loading and unloading, and an additional waiting time of 5~s was set after each loading--unloading cycle to allow the elastomer to fully recover its original shape. During the repeated tests, both the contact force and the output were measured. Assuming indoor use, the room temperature was controlled by air conditioning and set to 21.1~\textdegree{}C, 27.2~\textdegree{}C, and 31.2~\textdegree{}C.}
\begin{figure*}[!t]
    \centering
    \vspace{3mm}
    \includegraphics[width=\textwidth]{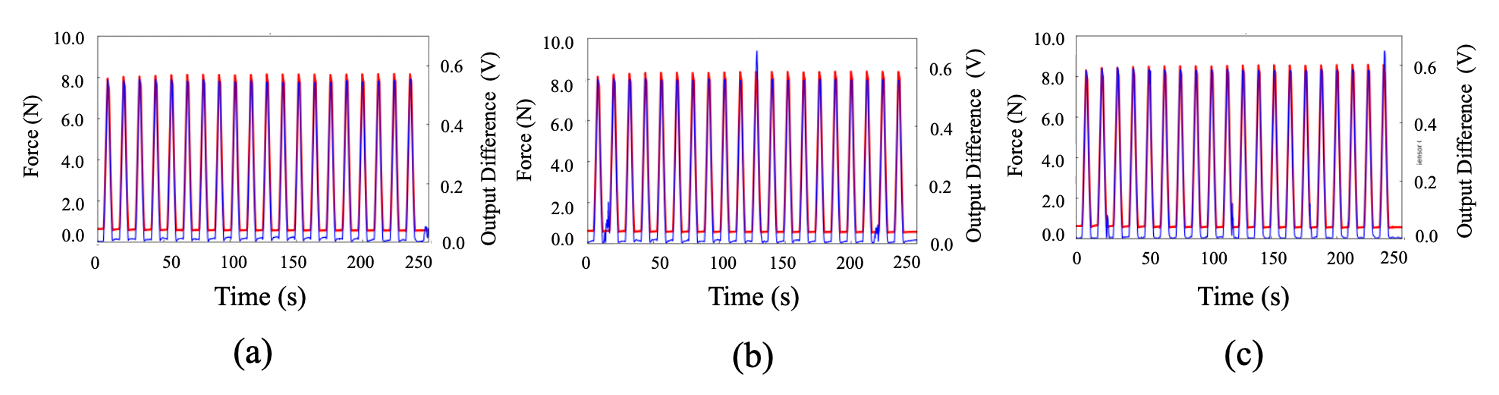}
    \vspace{-5mm}
    \caption{\textcolor{black}{Relationship between the applied force and the sensor output under different temperature conditions.
(a) 21.1~\textdegree{}C, (b) 27.2~\textdegree{}C, and (c) 31.2~\textdegree{}C. No variation in sensor output was observed under these temperature conditions, confirming that the sensor output remains stable in a room-temperature environment.
 }}
    \vspace{-6mm}
    \label{ex3}
\end{figure*}

\textcolor{black}{The experimental results are shown in Fig.~\ref{ex3}. Some trials showed an increase in the output. This may have been caused by the way the sensor was mounted. The results confirm that the sensor output did not vary with temperature. To quantitatively evaluate this result, the sensitivity and its standard deviation were derived. The sensitivities were 0.07294 $\pm$ 0.00129~V/N at 21.1~\textdegree{}C, 0.07164 $\pm$ 0.00116~V/N at 27.2~\textdegree{}C, and 0.07204 $\pm$ 0.00204~V/N at 31.2~\textdegree{}C, respectively. No substantial differences were observed in either the mean values or the standard deviations. These results indicate that the output does not change significantly under room-temperature conditions.}

\subsection{Soft Robot Sensor}
 \textcolor{black}{To verify the practicality of the proposed sensing method, a tactile-sensing module was mounted on a soft robotic gripper with a soft skin to examine whether the output changed during grasping.}

\begin{figure}[!t]
\centerline{\includegraphics[width=1\columnwidth]{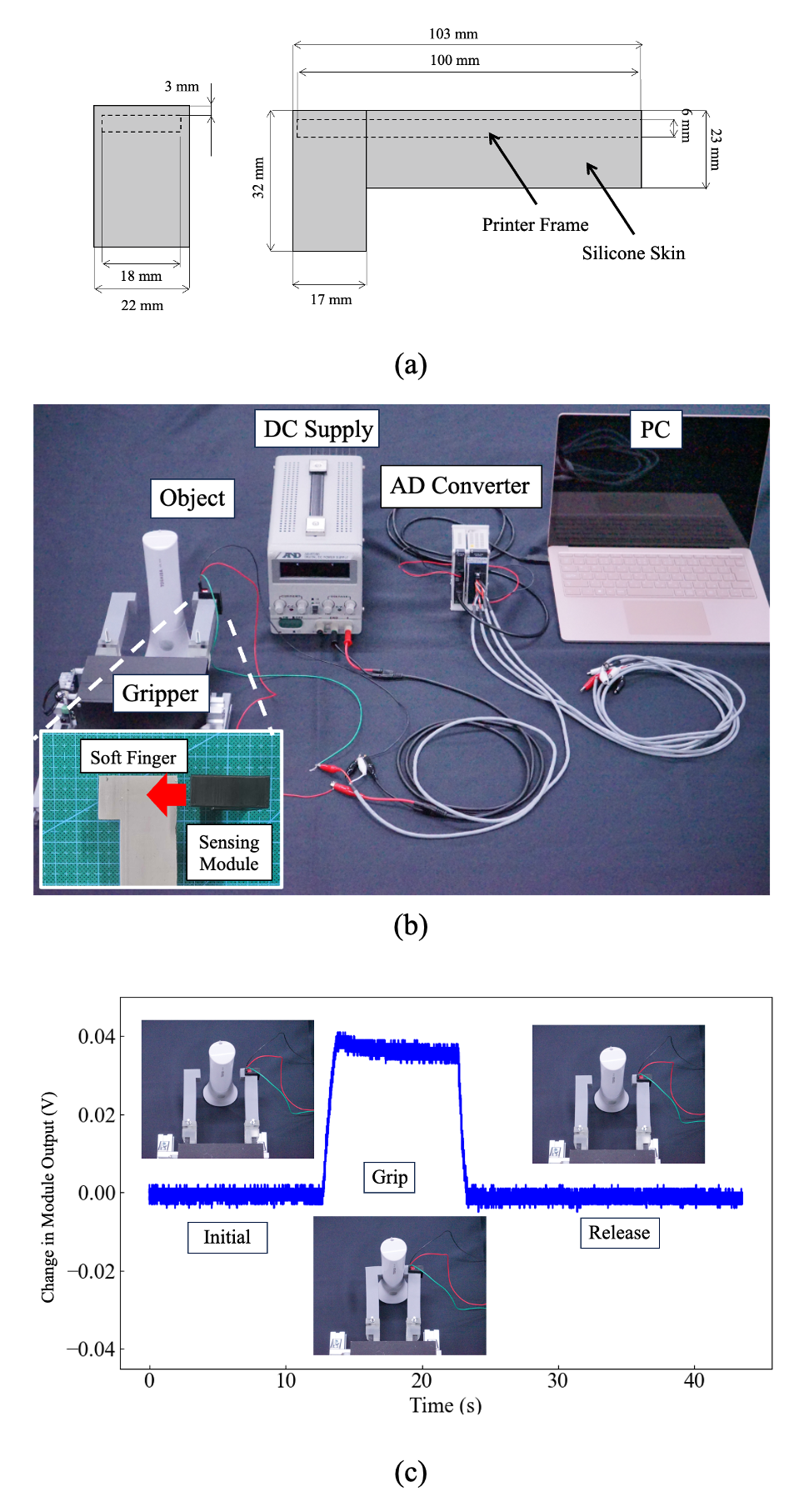}}
\caption{Experimental setup using a gripper and the corresponding results. (a) Dimensions of the fabricated soft finger. A resin frame was embedded inside, and the structure was covered with silicone rubber. Attaching the tactile-sensing module to the gripper created a soft robotic finger. (b) Overview of the experimental setup. The gripper was used to grasp a plastic rod. Signals from the tactile-sensing module were recorded on a PC, which also output control signals to the gripper. (c) Results of the grasping experiment. The object was grasped, and the output of the tactile-sensing module changed accordingly.}
\label{fig10}
\end{figure}

\textcolor{black}{The robotic fingers were fabricated as shown in Fig.~\ref{fig10}(a). The dashed section in the figure represents the resin frame that transmits force and is connected to a gripper. The frame was covered with silicone rubber, and the color was selected according to the results shown in Fig.~\ref{fig3}(b). The experimental setup used to evaluate the responses is illustrated in Fig.~\ref{fig10}(b). The gripper was constructed using a servomotor (XM430-W210-T, ROBOTIS), a linear guide (SSE2B8-130, MISUMI), a rack (RGEAPL1.0ST-70-A5-B10-C40-D10, MISUMI), and spur gears (GEABPS1.0-28-8-A-8, MISUMI). The finger section employed the structure shown in Fig.~\ref{fig10}(a). The tactile-sensing module embedded with a photoreflector was fixed to the fingertip of the robotic finger.}

 \textcolor{black}{The target object was a plastic rod. The signals from the PC were transmitted to the servomotor via a USB communication converter (U2D2, ROBOTIS) to operate the robotic gripper. During the experiments, the robotic gripper performed the grasping and releasing motions, and the output of the tactile-sensing module was recorded using a data logger (NR-500, KEYENCE) at a sampling frequency of 1~kHz.}

 \textcolor{black}{Fig.~\ref{fig10}(c) shows the change in the output of the tactile-sensing module during the grasping motion. During release and at rest, the changes in the output remained close to zero, whereas during grasping, the output increased. This experiment confirmed that the output changes during grasping. A quantitative evaluation of the grasping force and calibration are beyond the scope of this study and left for future research. For practical implementation of the proposed method in robots, the calibration of the relationship between the output change and grasping force, as well as the derivation of a calibration curve through simulation, is required. In this study, the tactile-sensing module was inserted into the silicone rubber without additional adhesives. It remained firmly attached during the experiment. For long-term applications, however, incorporating an interlocking design could further improve mechanical stability.}

\begin{table*}[t]
\centering
\caption{\textcolor{black}{Performance comparison with representative tactile sensors. Values are reported as stated in each reference; NR indicates not reported. Definitions and test conditions vary across studies.}}
\label{tab:comparison_quant}
\label{ext1}
\small
\resizebox{\textwidth}{!}{%
\begin{tabular}{l l c c c c c}
\hline
Work & Principle & Force Range & Sensitivity & Response & Hysteresis (as reported) & Repeatability \\
\hline
\textbf{This Work} & External Photoreflector + Skin Deformation
& $\sim$7 N@3 mm
& 0.07 V/N
& 1 kHz; lag 5--706~ms
& 6.50--7.88\%FS (Mean$\pm$SD, n=5)
& NR \\
\cite{cAndrea2021} & Optical Taxel Array (Embedded Pad)
& 0--15 N
& 0.018 V/N
& 500 Hz; 9.2~ms
& 5-6\% (Error)
& 1.32\% (Error) \\
\cite{usui2021} & Capacitive (Sealed Liquid)
& 0--1 N
& 0.169 pF/N
& 20 ms
& 6.28\%
& NR \\
\cite{hikarifiber1} & Optical Microfiber
& 0--10 N
& 5.4\%/N (0--1 N)
& 10 ms integration
& NR
& NR \\
\cite{seidenyouryousiki2} & 3-Axis Capacitive
& 0--190 kPa ($\sim$12 N)
& 14.22 /N
& Less than 40 ms
& NR
& NR \\
\cite{Park2019} & Piezoresistive Elastomer
& NR
& NR
& NR
& 6.8\% (Area-Based)
& Drift $\le$4\% \\
\hline
\end{tabular}
}
\end{table*}

\begin{table*}[t]
\centering
\caption{\textcolor{black}{Integration-oriented comparison for soft-robot deployment.}}
\label{tab:comparison_integration}
\label{ext2}
\small
\setlength{\tabcolsep}{4pt}
\renewcommand{\arraystretch}{1.0}
\begin{tabular}{l c c c c c}
\hline
Work & Off-Interface & Embed in Skin & Replaceable & Wiring & Spatial Capability \\
\hline
\textbf{This Work} & Yes & No & Yes & Low & Single-Point (Scalable) \\
\cite{cAndrea2021} & No & Yes & No & High (PCB) & Array \\
\cite{usui2021} & Yes & Yes & Partial & Low & Single-Point \\
\cite{hikarifiber1} & Yes & Yes & Partial & Low & Single-Point \\
\cite{seidenyouryousiki2} & No & Yes & No & High & 3-axis point \\
\cite{Park2019} & No & No & No & Med & Array \\
\hline
\end{tabular}
\end{table*}

\section{CONCLUSION AND DISCUSSION}
 \textcolor{black}{We proposed a tactile-sensing method for soft robots that attaches an external module to the soft skin, eliminating embedded elements, reducing damage risk, and simplifying fabrication. A prototype module demonstrated contact-force sensing and the feasibility of integration with a soft gripper.}

 \textcolor{black}{First, we confirmed that the output caused by the contact force can be predicted using the proposed modeling approach, as shown in Fig.~\ref{fig7}(b). In the simulations, output estimation was performed according to the deformation values obtained through the finite-element analysis. If the deformation caused by the external forces can be predicted during the design phase, the expected output can be assessed in advance. The finite-element model used in this study served as an initial design tool. Further refinement would improve the predictive accuracy.}

 \textcolor{black}{Then, the proposed sensing method exhibited good responses and repeatability to an external force. In tactile sensors that use flexible conductive elastomer materials as sensing elements, the material viscosity can affect the response time and hysteresis characteristics. In contrast, the proposed method estimates the contact force by capturing the deformation of the soft robot skin without using specialized elastomer materials. However, this implies that the performance of the sensor can vary significantly depending on the mechanical properties of the skin material. For example, when highly viscous self-healing polymers with specific molecular structures are used~\cite{kosaka2025}, degradation of the response time and hysteresis characteristics is expected. Adjusting the properties of the material offers the possibility of tuning the sensor's performance. For example, the selection of a material with appropriate stiffness can alter the sensor's sensitivity. Using a softer material results in larger deformations under small forces, improving the sensitivity to contact force. Conversely, a stiffer material requires larger forces to produce measurable deformations, enabling a wider dynamic range for the sensor. Thus, careful selection of the materials used in constructing a soft robot can facilitate the design and optimization of the sensing characteristics for specific applications.}

 \textcolor{black}{Finally, application of the tactile-sensing method to soft robotic fingers confirmed that the contact could be successfully detected. The proposed method suggests the possibility of obtaining sensory information without embedding sensors or special materials in the soft skin of the robot. Although the present study focused on tactile information in the form of the contact force, the method can also be applied to other types of somatosensory information. For example, similar to human skin, joint bending in a soft robot can cause characteristic deformations of the surrounding skin~\cite{Mia2025}. According to this concept, measuring skin deformations near the joints could enable estimation of joint angles. Similarly, this information can be obtained if the internal state of an actuator produces measurable deformations on the skin surface. Therefore, the applicability of the proposed method depends heavily on the mechanical design of the soft robots. We employed a photoreflector to convert the deformation caused by contact force into a change in the received light intensity. However, any sensing principle capable of accurately capturing the deformation can be used. For example, methods that utilize acoustic reflection~\cite{monica2025} or optical waveguides~\cite{yamamoto2024} can be applied for force estimation within this framework.}

 \textcolor{black}{In preliminary trials with the prototype soft robot shown in Fig.~\ref{fig10}, when the finger lacked an internal structure analogous to bones, no skin deformation occurred during grasping, and the contact could not be detected. However, if the deformation of a soft robot's skin under external forces can be increased through an appropriate design, sensing should be possible without embedding rigid components. An appropriate mechanical design that ensures motion-induced skin deformations are measurable would enable the acquisition of robot state information. Incorporating design methodologies such as topology optimization can further enhance the deformation characteristics, enabling more effective soft robot designs for the proposed sensing approach.}

\textcolor{black}{The tactile-sensing module embeds a single photoreflector; therefore, it primarily estimates the net force applied to the entire contact area rather than the spatial distribution of pressure. Spatial information can be obtained by placing multiple sensing modules at different locations on the robot's surface. Such multi-site force measurements can be used for downstream tasks reported in prior soft-robotics studies~\cite{Zuo2021, Shorthose2022, Shorthose2025}, including grasp-state estimation and object identification.}
\textcolor{black}{To increase the number of sensing sites, the module should be miniaturized. In the current prototype, the thickness is constrained by the photoreflector placement and the optical/mechanical stack, limiting the reduction achievable through thinning the ABS housing. Future miniaturization requires a custom optical design, including the selection and direct implementation of smaller emitters and detectors. If the optical stack is reduced, multiple optical sensing elements can be embedded within a single module, enabling local force features to be extracted and potentially providing spatially resolved contact information.}

\textcolor{black}{Table~\ref{ext1} summarizes the performance comparison~\cite{cAndrea2021, usui2021, hikarifiber1, seidenyouryousiki2, Park2019}. Although this study does not primarily aim to maximize sensing performance, the comparison indicates that our method achieves hysteresis within the range reported for representative soft tactile sensors. 
Compared with a tactile sensor that embed photoreflectors directly beneath the loading region~\cite{cAndrea2021}, our approach exhibits a higher sensitivity. This result contradicts the intuitive expectation that the embedded type would be more sensitive, suggesting that sensitivity can be improved by varying materials and structural design.
Table~\ref{ext2} provides an integration-oriented comparison. Existing approaches typically require embedding sensing components within the soft skin, which can complicate fabrication and hinder replacement. In contrast, by adopting an externally attachable and non-embedded design, we enable tactile sensing for soft robots with simplified fabrication and straightforward module replacement. Note that the quantitative metrics in Tables~\ref{ext1} and~\ref{ext2} are reported as stated in the original references; definitions and test conditions may differ across studies.}

\textcolor{black}{Future work includes full calibration to map voltage to force across indentation speeds, durability testing, and long-term use while implementing the module in a soft robot, as well as multi-module arrays for shear and torque estimation. We will investigate better design and placement of the modules by solving these topics.  }

\section{ACKNOWLEDGMENT}
\textcolor{black}{The authors thank Prof.\ Hiroyuki Kajimoto (The University of Electro-Communications) for valuable advice on the simulations.}

\section{DECLARATION OF COMPETING INTEREST}
\textcolor{black}{The authors declare no conflict of interest that could influence the findings presented in this document.}

\section{DATA AVAILABILITY}
\textcolor{black}{Data will be made available upon reasonable request.}


\begin{thebibliography}{10}
\providecommand{\url}[1]{#1}
\csname url@samestyle\endcsname
\providecommand{\newblock}{\relax}
\providecommand{\bibinfo}[2]{#2}
\providecommand{\BIBentrySTDinterwordspacing}{\spaceskip=0pt\relax}
\providecommand{\BIBentryALTinterwordstretchfactor}{4}
\providecommand{\BIBentryALTinterwordspacing}{\spaceskip=\fontdimen2\font plus
\BIBentryALTinterwordstretchfactor\fontdimen3\font minus \fontdimen4\font\relax}
\providecommand{\BIBforeignlanguage}[2]{{%
\expandafter\ifx\csname l@#1\endcsname\relax
\typeout{** WARNING: IEEEtran.bst: No hyphenation pattern has been}%
\typeout{** loaded for the language `#1'. Using the pattern for}%
\typeout{** the default language instead.}%
\else
\language=\csname l@#1\endcsname
\fi
#2}}
\providecommand{\BIBdecl}{\relax}
\BIBdecl

\bibitem{katarobomondai}
\textcolor{black}{D.~Trivedi, C.~D. Rahn, W.~M. Kier, and I.~D. Walker, ``Soft robotics: Biological inspiration, state of the art, and future research,'' \emph{Applied Bionics and Biomechanics}, vol.~5, no.~3, pp. 99--117, 2008.}

\bibitem{gendairobot}
\textcolor{black}{F.~Tanaka, A.~Cicourel, and J.~R. Movellan, ``Socialization between toddlers and robots at an early childhood education center,'' \emph{Proceedings of the National Academy of Sciences}, vol. 104, no.~46, pp. 17\,954--17\,958, 2007.}

\bibitem{gendairobot2}
\textcolor{black}{C.~D. Kidd and C.~Breazeal, ``Robots at home: Understanding long-term human-robot interaction,'' in \emph{Proceedings of 2008 IEEE/RSJ International Conference on Intelligent Robots and Systems (IROS)}, 2008, pp. 3230--3235.}

\bibitem{Nakadegawa2017}
\textcolor{black}{T.~Nakadegawa, H.~Ishizuka, and N.~Miki, ``Three-axis scanning force sensor with liquid metal electrodes,'' \emph{Sensors and Actuators A: Physical}, vol. 264, pp. 260--267, 2017.}

\bibitem{Ohashi2023}
\textcolor{black}{H.~Ohashi, T.~Yasuda, T.~Kawasetsu, and K.~Hosoda, ``Soft tactile sensors having two channels with different slopes for contact position and pressure estimation,'' \emph{IEEE Sensors Letters}, vol.~7, no.~5, pp. 1--4, 2023.}

\bibitem{Hirai2021}
\textcolor{black}{S.~Hirai, T.~Nagatomo, T.~Hiraki, H.~Ishizuka, Y.~Kawahara, and N.~Miki, ``Micro elastic pouch motors: Elastically deformable and miniaturized soft actuators using liquid-to-gas phase change,'' \emph{IEEE Robotics and Automation Letters}, vol.~6, no.~3, pp. 5373--5380, 2021.}

\bibitem{Tanaka2025}
\textcolor{black}{H.~Tanaka, H.~Hirai, and K.~Hosoda, ``Position control of mckibben-type pneumatic artificial muscle via port-hamiltonian approach,'' \emph{IEEE Robotics and Automation Letters}, vol.~10, no.~6, pp. 6384--6391, 2025.}

\bibitem{GarradSoftMatterComputer2019}
\textcolor{black}{M.~Garrad, G.~Soter, A.~T. Conn, H.~Hauser, and J.~Rossiter, ``A soft matter computer for soft robots,'' \emph{Science Robotics}, vol.~4, no.~33, p. eaaw6060, 2019.}

\bibitem{Hashimoto2023}
\textcolor{black}{Y.~Hashimoto, H.~Ishizuka, T.~Kawasetsu, T.~Horii, S.~Ikeda, and O.~Oshiro, ``Selective voltage application to connected loads using soft matter computer based on conductive droplet interval design,'' \emph{IEEE Robotics and Automation Letters}, vol.~8, no.~3, pp. 1747--1754, 2023.}

\bibitem{atudenkoukasiki}
\textcolor{black}{W.~Liu, P.~Yu, C.~Gu, X.~Cheng, and X.~Fu, ``Fingertip piezoelectric tactile sensor array for roughness encoding under varying scanning velocity,'' \emph{IEEE Sensors Journal}, vol.~17, no.~21, pp. 6867--6879, 2017.}

\bibitem{atudenkoukasikiondo}
\textcolor{black}{L.~Persano, C.~Dagdeviren, Y.~Su, Y.~Zhang, S.~Girardo, D.~Pisignano, Y.~Huang, and J.~A. Rogers, ``High performance piezoelectric devices based on aligned arrays of nanofibers of poly(vinylidenefluoride-co-trifluoroethylene),'' \emph{Nature Communications}, vol.~4, p. Art. no. 1633, 2013, article number: 1633.}

\bibitem{zikisiki}
\textcolor{black}{M.~Rehan, M.~M. Saleem, M.~I. Tiwana, R.~I. Shakoor, and R.~Cheung, ``A soft multi-axis high force range magnetic tactile sensor for force feedback in robotic surgical systems,'' \emph{Sensors}, vol.~22, no.~9, p. 3500, 2022.}

\bibitem{teikousiki1}
\textcolor{black}{Y.~Jung, K.~K. Jung, D.~H. Kim, D.~H. Kwak, and J.~S. Ko, ``Linearly sensitive and flexible pressure sensor based on porous carbon nanotube/polydimethylsiloxane composite structure,'' \emph{Polymers}, vol.~12, no.~7, p. 1499, 2020.}

\bibitem{teikousiki2}
\textcolor{black}{M.~Shimojo, A.~Namiki, M.~Ishikawa, R.~Makino, and K.~Mabuchi, ``A tactile sensor sheet using pressure conductive rubber with electrical-wires stitched method,'' \emph{IEEE Sensors Journal}, vol.~4, no.~5, pp. 589--596, 2004.}

\bibitem{seidenyouryousiki1}
\textcolor{black}{Y.~Liu, H.~Wo, S.~Huang, Y.~Huo, H.~Xu, S.~Zhan, M.~Li, X.~Zeng, H.~Jin, L.~Zhang, X.~Wang, S.~Dong, J.~Luo, and J.~M. Kim, ``A flexible capacitive 3d tactile sensor with cross-shaped capacitor plate pair and composite structure dielectric,'' \emph{IEEE Sensors Journal}, vol.~21, no.~2, pp. 1378--1385, 2021.}

\bibitem{seidenyouryousiki2}
\textcolor{black}{L.~Viry, A.~Levi, M.~Totaro, A.~Mondini, V.~Mattoli, B.~Mazzolai, and L.~Beccai, ``Flexible three-axial force sensor for soft and highly sensitive artificial touch,'' \emph{Advanced Materials}, vol.~26, no.~17, pp. 2659--2664, 2014.}

\bibitem{seidenyouryousiki3}
\textcolor{black}{H.~Zhang and M.~Y. Wang, ``Multi-axis soft sensors based on dielectric elastomer,'' \emph{Soft Robotics}, vol.~3, no.~1, pp. 3--12, 2016.}

\bibitem{hikarifiber1}
\textcolor{black}{C.~Jiang, Z.~Zhang, J.~Pan, Y.~Wang, L.~Zhang, and L.~Tong, ``Finger-skin-inspired flexible optical sensor for force sensing and slip detection in robotic grasping,'' \emph{Advanced Materials Technologies}, vol.~6, no.~10, pp. 2\,100\,285--2\,100\,294, 2021.}

\bibitem{hikarifiber2}
\textcolor{black}{Y.~Du, Q.~Yang, and J.~Huang, ``Soft prosthetic forefinger tactile sensing via a string of intact single mode optical fiber,'' \emph{IEEE Sensors Journal}, vol.~17, no.~22, pp. 7455--7459, 2017.}

\bibitem{hikarifiber3}
\textcolor{black}{R.~Ahmadi, M.~Packirisamy, J.~Dargahi, and R.~Cecere, ``Discretely loaded beam-type optical fiber tactile sensor for tissue manipulation and palpation in minimally invasive robotic surgery,'' \emph{IEEE Sensors Journal}, vol.~12, no.~1, pp. 22--32, 2012.}

\bibitem{umekomi1}
\textcolor{black}{R.~P. Rocha, P.~A. Lopes, A.~T.~d. Almeida, M.~Tavakoli, and C.~Majidi, ``Fabrication and characterization of bending and pressure sensors for a soft prosthetic hand,'' \emph{Journal of Micromechanics and Microengineering}, vol.~5, no.~2, pp. 034\,001--034\,010, 2018.}

\bibitem{umekomi2}
\textcolor{black}{R.~L. Truby, R.~K. Katzschmann, J.~A. Lewis, and D.~Rus, ``Soft robotic fingers with embedded ionogel sensors and discrete actuation modes for somatosensitive manipulation,'' in \emph{Proceedings of 2019 2nd IEEE International Conference on Soft Robotics (RoboSoft)}, 2019, pp. 322--329.}

\bibitem{umekomi3}
\textcolor{black}{Y.~Yang and Y.~Chen, ``Innovative design of embedded pressure and position sensors for soft actuators,'' \emph{IEEE Robotics and Automation Letters}, vol.~3, no.~2, pp. 656--663, 2018.}

\bibitem{dansen}
\textcolor{black}{A.~Gruebele, J.-P. Roberge, A.~Zerbe, W.~Ruotolo, T.~M. Huh, and M.~R. Cutkosky, ``A stretchable capacitive sensory skin for exploring cluttered environments,'' \emph{IEEE Robotics and Automation Letters}, vol.~5, no.~2, pp. 1750--1757, 2020.}

\bibitem{Mia2025}
\textcolor{black}{M.~Rupani, L.~D. Cleland, and H.~P. Saal, ``Local postural changes elicit extensive and diverse skin stretch around joints, on the trunk and the face,'' \emph{Journal of The Royal Society Interface}, vol.~22, no. 223, p. 20240794, 2025.}

\bibitem{johansson2004}
\textcolor{black}{R.~S. Johansson and I.~Birznieks, ``First spikes in ensembles of human tactile afferents code complex spatial fingertip events,'' \emph{Nature neuroscience}, vol.~7, no.~2, pp. 170--177, 2004.}

\bibitem{soter2019}
\textcolor{black}{G.~Soter, M.~Garrad, A.~T. Conn, H.~Hauser, and J.~Rossiter, ``Skinflow: A soft robotic skin based on fluidic transmission,'' in \emph{Proceedings of 2019 2nd IEEE International Conference on Soft Robotics (RoboSoft)}, 2019, pp. 355--360.}

\bibitem{usui2021}
\textcolor{black}{T.~Usui, H.~Ishizuka, T.~Kawasetsu, K.~Hosoda, S.~Ikeda, and O.~Oshiro, ``Soft capacitive tactile sensor using displacement of air-water interface,'' \emph{Sensors and Actuators A: Physical}, vol. 332, no.~1, p. 113133, 2021.}

\bibitem{hamaguchi2021}
\textcolor{black}{S.~Hamaguchi, T.~Kawasetsu, T.~Horii, H.~Ishihara, R.~Niiyama, K.~Hosoda, and M.~Asada, ``Soft inductive tactile sensor using flow-channel enclosing liquid metal,'' \emph{IEEE Robotics and Automation Letters}, vol.~5, no.~3, pp. 4028--4034, 2020.}

\bibitem{Martin2022}
\textcolor{black}{Y.~J. Martin, D.~Bruder, and R.~J. Wood, ``A proprioceptive method for soft robots using inertial measurement units,'' in \emph{2022 IEEE/RSJ International Conference on Intelligent Robots and Systems (IROS)}, 2022, pp. 9379--9384.}

\bibitem{nakatani2011}
\textcolor{black}{M.~Nakatani, T.~Kawasoe, K.~Shiojima, K.~Koketsu, S.~Kinoshita, and J.~Wada, ``Wearable contact force sensor system based on fingerpad deformation,'' in \emph{Proceedings of 2011 IEEE World Haptics Conference}, 2011, pp. 323--328.}

\bibitem{Saito2019}
\textcolor{black}{A.~Saito, W.~Kuno, W.~Kawai, N.~Miyata, and Y.~Sugiura, ``Estimation of fingertip contact force by measuring skin deformation and posture with photo-reflective sensors,'' in \emph{Proceedings of the 10th Augmented Human International Conference 2019}, 2019, pp. 1--6.}

\bibitem{kaneko2025}
\textcolor{black}{S.~Kaneko, H.~Ishizuka, H.~Yoshimura, and H.~Kajimoto, ``Utilization of skin color change for image-based tactile sensing,'' \emph{Medical Engineering \& Physics}, vol. 140, p. 104357, 2025.}

\bibitem{ohmura2006}
\textcolor{black}{Y.~Ohmura, Y.~Kuniyoshi, and A.~Nagakubo, ``Conformable and scalable tactile sensor skin for curved surfaces,'' in \emph{Proceedings 2006 IEEE International Conference on Robotics and Automation (ICRA)}, 2006, pp. 1348--1353.}

\bibitem{cAndrea2021}
\textcolor{black}{A.~Cirillo, M.~Costanzo, G.~Laudante, and S.~Pirozzi, ``Tactile sensors for parallel grippers: Design and characterization,'' \emph{Sensors}, vol.~21, no.~5, p. 1915, 2021.}

\bibitem{hayase2023}
\textcolor{black}{G.~Hayase, ``Optical tactile sensor using scattering inside sol-gel-derived flexible macroporous monoliths,'' \emph{Sensors and Actuators A: Physical}, vol. 354, p. 114253, 2023.}

\bibitem{niu2022}
\textcolor{black}{L.~Niu, Y.~Su, H.~Yang, G.~Liu, H.~Gao, and Z.~Deng, ``Dynamic finite element modeling and simulation of soft robots,'' \emph{Chinese Journal of Mechanical Engineering}, vol.~35, 12 2022.}

\bibitem{qiu2019}
\textcolor{black}{J.~Qiu, X.~Guo, R.~Chu, S.~Wang, W.~Zeng, L.~Qu, Y.~Zhao, F.~Yan, and G.~Xing, ``Rapid-response, low detection limit, and high-sensitivity capacitive flexible tactile sensor based on three-dimensional porous dielectric layer for wearable electronic skin,'' \emph{ACS Applied Materials \& Interfaces}, vol.~11, no.~43, pp. 40\,716--40\,725, 2019.}

\bibitem{Wang2017}
\textcolor{black}{H. Wang, J. Kow, N. Raske, D. Boer, M. Ghajari, R. Hewson, A. Alazmani, and P. Culmer, ``Robust and high-performance soft inductive tactile sensors based on the Eddy-current effect,'' \emph{Sensors and Actuators A: Physical}, vol. 271, pp. 44-52, 2017.}

\bibitem{kosaka2025}
\textcolor{black}{S.~Kosaka, K.~Kimura, S.~Yamamoto, H.~Ishizuka, Y.~Masuda, P.~Punpongsanon, S.~Ikeda, and O.~Oshiro, ``Reconfigurable soft pneumatic actuators using multi-material self-healing polymers,'' \emph{IEEE Robotics and Automation Letters}, vol.~10, no.~5, pp. 4938--4945, 2025.}

\bibitem{monica2025}
\textcolor{black}{M.~S. Li and H.~S. Stuart, ``Acoustac: Tactile sensing with acoustic resonance for electronics-free soft skin,'' \emph{Soft Robotics}, vol.~12, no.~1, pp. 109--123, 2025.}

\bibitem{yamamoto2024}
\textcolor{black}{S.~Yamamoto, H.~Ishizuka, S.~Ikeda, and O.~Oshiro, ``A self-healing tactile sensor using an optical waveguide,'' in \emph{Proceedings of Asia Haptics 2024}, 2024, pp. 1--3.}

\bibitem{Zuo2021}
\textcolor{black}{R.~Zuo, Z.~Zhou, B.~Ying, and X.~Liu, ``A Soft Robotic Gripper with Anti-Freezing Ionic Hydrogel-Based Sensors for Learning-Based Object Recognition,'' in \emph{Proceedings of 2021 IEEE International Conference on Robotics and Automation (ICRA)}, 2021, pp. 12164--12169.}

\bibitem{Shorthose2022}
\textcolor{black}{O. Shorthose, A. Albini, L. He, and P. Maiolino, ``Design of a 3D-Printed Soft Robotic Hand With Integrated Distributed Tactile Sensing,'' \emph{IEEE Robotics and Automation Letters}, vol.~7, no.~2, pp. 3945--3952, 2022.}

\bibitem{Shorthose2025}
\textcolor{black}{O. Shorthose, L. Scimeca, A. Albini, and P. Maiolino, ``Bioinspired Tactile Object Identification Leveraging Deep
Learning and Soft Body Compliance,'' \emph{Advanced Intelligent Systems}, vol. 7, p. 2400802, 2025.}

\bibitem{Park2019}
\textcolor{black}{K. Park, S. Kim, H. Lee, I. Park, and J. Kim, ``Low-hysteresis and low-interference soft tactile sensor using aconductive coated porous elastomer and a structure forinterference reduction,'' \emph{Sensors and Actuators A: Physical}, vol. 295, pp. 541-550, 2019.}

\end{thebibliography}

\end{document}